\documentclass[journal]{IEEEtran}
\IEEEoverridecommandlockouts
\usepackage{cite}
\usepackage{amsmath,amssymb,amsfonts}
\usepackage[algo2e]{algorithm2e}
\usepackage{graphicx}
\usepackage{textcomp}
\usepackage{xcolor}
\usepackage{amsmath}
\usepackage{caption}
\usepackage{subcaption}
\usepackage{enumitem}

\newcommand{\RomanNumeralCaps}[1]{\MakeUppercase{\romannumeral #1}}

\SetKwInput{KwInput}{Input}                
\SetKwInput{KwOutput}{Output} 
\usepackage[utf8]{inputenc}
\usepackage[english]{babel}
\newtheorem{theorem}{Theorem}
\usepackage{algorithm,algorithmic,amsmath}
\usepackage{mathtools}
\usepackage{amsmath}
\usepackage{comment}
\def\BibTeX{{\rm B\kern-.05em{\sc i\kern-.025em b}\kern-.08em
    T\kern-.1667em\lower.7ex\hbox{E}\kern-.125emX}}
\usepackage{optidef}

\graphicspath{{./figs/}}

\begin{document}

\title{A General Framework for Uncertainty Quantification via Neural SDE-RNN\\
}

\author{
Shweta Dahale, \IEEEmembership{Graduate Student Member,~IEEE}, Sai Munikoti, \IEEEmembership{Member,~IEEE}, and 
\\ Balasubramaniam Natarajan, \IEEEmembership{Senior Member,~IEEE} \\

\thanks{S. Dahale and B. Natarajan are with the Electrical and Computer Engineering, Kansas State University, Manhattan, KS-66506, USA (e-mail: sddahale@ksu.edu, bala@ksu.edu). S.Munikoti is with the Data Science and Machine Intelligence group, Pacific Northwest National Laboratory, USA, (e-mail: sai.munikoti@pnnl.gov). 

This material is based upon work supported by the Department  of  Energy,  Office  of  Energy  Efficiency  and  Renewable Energy  (EERE),  Solar  Energy  Technologies  Office,  under Award Number DE-EE0008767.

}}

\maketitle

\begin{abstract}
Uncertainty quantification is a critical yet unsolved challenge for deep learning, especially for the time series imputation with irregularly sampled measurements. To tackle this problem, we propose a novel framework based on the principles of recurrent neural networks and neural stochastic differential equations for reconciling irregularly sampled measurements. We impute measurements at any arbitrary timescale and quantify the uncertainty in the imputations in a principled manner.  Specifically, we derive analytical expressions for quantifying and propagating the epistemic and aleatoric uncertainty across  time instants. Our experiments on the IEEE 37 bus test distribution system reveal that our framework can outperform state-of-the-art uncertainty quantification approaches for time-series data imputations.
\end{abstract}


\section{Introduction}


Irregularly sampled multivariate time-series data are found in several domains spanning from bio-medical systems to the smart grid. This is primarily the outcome of two factors. The first source is the inherent system design where data generation happens at different rate. For instance, in a smart grid, different sensors sense and transmit measurements at different time scales, i.e., advanced metering infrastructure (AMI) samples at 15-min interval, data acquisition (SCADA) sensors at 1-min interval, etc.\cite{dahale2022recursive}.
The second factor corresponds to system malfunctions that causes frequent unavailability of data samples. For example, communication impairments in the smart grid result in loss of data, further aggravating this issue \cite{dahale2022bayesian}. It is important to reconcile such multi-scale intermittent measurements at a common timescale in order to enhance the situational awareness of the grid. Typically, this kind of irregularly sampled data is first imputed (i.e., estimate missing values) to a uniform scale so that an effective analysis can be carried out with the processed data. To this end, several methods are being proposed in the literature such as linear interpolation \cite{gomez2014state}, kNN \cite{al2016state} and multitask Gaussian process \cite{dahale2022recursive,9637824}. Recently, neural ODES (NeuODES) have shown to be a very effective framework for handling irregular sampled data \cite{chen2018neural, dahale2022latent, 9180216, lu2021neural}. This is primarily due to the fact that NeuODES can estimate values in a continuous space (using dynamics) unlike discrete space (proportional to the number of layers) in a vanilla neural network.  

However, most of the data imputation frameworks solely offer a point estimate of the missing value, ignoring their uncertainties.
It is crucial to provide confidence scores with prediction so that an informed decision can be made while deployed in the real-world applications. Uncertainty quantification (UQ) of the imputed measurements can be helpful in various ways, including (i) variance-informed state estimation where variance (confidence interval) of the imputed measurement's can be used to modulate the inputs of the estimator \cite{dahale2022bayesian}. For instance, high variance input is disregarded or corrected before being fed to the estimator.
(ii) Effective data sampling where UQ helps in selecting appropriate measurements from high or low-fidelity sensors.

Uncertainty in deep neural network-based models such as NeuODES arises due to two components. The first factor is with respect to the model parameters and it arises due to the model either being blind to some portion of the data distribution which has not been seen while it's training or due to its over-parameterization. There are various ways to handle this and can be quantified in a computationally efficient manner. On the other hand, the second class of uncertainty is related to the input data, also known as aleatoric uncertainty. It is inherent in nature and cannot be alleviated by any means of model engineering \cite{gal2016dropout}. Modeling and quantifying these uncertainties are critical with regard to time series imputation since the sensors are often associated with noise, and one cannot cover the entire data distribution while training the models. 

There exist a few works in the literature that quantify uncertainty in NeuODES-based frameworks, \cite{li2020scalable, jia2019neural, hegde2018deep,herrera2020neural}. Authors in \cite{herrera2020neural} proposed a neural Jump ODE framework for modeling the conditional expectation of a stochastic process in an irregularly observed time series. 
However, most of these works only retrieve the epistemic part and ignore the aleatoric part. It is important to incorporate the aleatoric part, especially for critical complex systems where sensor-generated data is pivotal in various operational decision-making. Therefore, we propose a novel framework to quantify the uncertainties in a comprehensive manner. In this regard, we first formulate an SDE-RNN framework that combines the stochastic  differential equation (SDE) form of NeuODES and recurrent neural network (RNN) for imputation. Specifically, SDE is chosen since it can propagate both mean (prediction) and variances (uncertainty) along the NeuODES framework in an efficient manner. Furthermore, RNN helps us to capture the inputs at observed time instances, and effectively carry forward the information to future states via its memory cell. Altogether, their combination offers an powerful and efficient approach to quantify and propagate both uncertainties from the input to final predictions.



The main contributions of this paper are summarized below:
\begin{itemize}
    \item We formulate a novel SDE-RNN which combines the principle of stochastic differential equations and neural networks to model the irregularly sampled time series.
    \item The proposed SDE-RNN approach allows us to quantify both aleatoric and epistemic uncertainty in the imputed measurements. Analytical results capturing these sources of uncertainty have been derived. We derive the theoretical expressions of uncertainty propagation for Gated Recurrent Unit (GRU) model.
    \item Simulation results on the power distribution test system (IEEE 37 network) demonstrate the effectiveness of the proposed approach compared to the classic baseline (RNN + MC dropout).
\end{itemize}

The paper is organized as follows. Section \RomanNumeralCaps{2} presents a background of NeuODES with a literature review. Section \RomanNumeralCaps{3} formulates our fundamental SDE-RNN framework 
followed by the detailed process of uncertainty quantification in 
Section \RomanNumeralCaps{4}. Experiments are discussed in Section \RomanNumeralCaps{5} with conclusions and future work in Section \RomanNumeralCaps{6}.

\section{Background and Related work}
This section provides an overview of the UQ literature for unevenly sampled time series imputation models. Additionally, the mathematical background of neural ODES, which serves as a fundamental module in our proposed approach, is provided. 
\subsection{Related work}
Recurrent neural networks (RNN) form the first choice for modeling high dimensional, regularly sampled time series data. The success of RNNs is due to its memory and cell state modules that can capture long range dependencies. A natural extension of
RNNs to unevenly sampled time series data is to divide the
timeline into equally-sized intervals \cite{lipton2016directly}, \cite{che2018recurrent}, and impute or aggregate
observations using averages. This pre-processing stage 
maligns the information, particularly about the timing of
measurements, as they constitute a piece of important information about the data. Another approach uses a simple exponential decay between observations, and updates the hidden states accordingly \cite{che2018recurrent}. But the states of RNN-decay model approach zero if the time gaps between observations are high. 
However, Neural ODE, which is a novel framework 
combining deep learning principles and differential equations has been found to be suitable for modeling irregularly sampled time series \cite{chen2018neural}. Neural ODEs can systematically accommodate continuous time series data using a technique that captures the continuous nature of hidden states. Recently, \cite{rubanova2019latent} combines  neural ODEs with RNN, leading to a method referred to as ODE-RNN. The hidden states in these ODE-RNN models obey an ODE between consecutive observations and are only updated at observed time instances. The ODE-RNN models are suitable for sparse observations and thus very effective for imputation and one-step prediction.  Authors in \cite{li2020scalable} propose a Latent SDE with an adjoint method for continuous time series modeling. Neural SDEs driven by stochastic processes with jumps is proposed in \cite{jia2019neural} to learn continuous and discrete dynamic behavior. Generative models built from SDEs whose drift
and diffusion coefficients are samples from a Gaussian process are introduced in \cite{hegde2018deep}. A neural Jump ODE framework for modeling the conditional expectation of a stochastic process in an irregularly observed time series is proposed in \cite{herrera2020neural}. However, all neural ODE driven methodologies only offer a point estimate and fail to quantify the uncertainty associated with the estimates. 

UQ approaches in general neural networks are mainly categorized into two classes: Bayesian and Ensemble methods. Bayesian approaches, such as Bayesian neural nets proposed in \cite{denker1990transforming} quantify the uncertainty by imposing probability distributions over model parameters. Though, these approaches provide a principled way of uncertainty quantification, the exact inference of the parameter posteriors is intractable. Also, specifying the priors for the parameters of deep neural networks becomes challenging when size of the network increases. To deal with these challenges, approximation methods like variational inference \cite{blundell2015weight}, Laplace approximation \cite{ritter2018scalable}, Assumed density filtering \cite{munikoti2023general} and stochastic gradient MCMC \cite{li2016preconditioned} are used. Authors in \cite{gal2016dropout} proposed to use Monte-Carlo Dropout (MC-dropout) during inference to quantify the uncertainty in neural networks. However, MC dropout is computationally expensive for larger network and only capture epistemic uncertainty. Non-Bayesian approaches have also been used for UQ but they also demand a large computational effort \cite{lakshminarayanan2017simple}. A different route has been taken in \cite{kong2020sde} where SDEs are used to quantify both aleatoric and epistemic uncertainty by training the SDE model based on out-of-distribution (OOD) training data. However, 
this approach fails to evaluate uncertainty in a principled manner.

\subsection{Background: Neural ODE}
Consider time-series sensor data $\{x_i, t_i\}_{i=0}^{N}$, where the measurements $\{ x_i\}_{i=0}^{N} \in \mathbb{R}^d$ are obtained at times  $\{ t_i\}_{i=0}^{N}$. 
The goal of timeseries imputation is to reconcile the unevenly sampled  measurements at at the finest time resolution.
Neural ODE exploits the continuous-time dynamics of variables from input state to final predictions, unlike a standard deep neural network which only performs a limited number of transformations depending on the number of layers chosen in the architecture. The state transitions in RNNs are generalized to continuous time dynamics specified by a neural network. We denote this model as ODE-RNN. The hidden state in ODE-RNN is a solution to an ODE initial-value problem, given as:
\begin{equation}
    \frac{d\mathbf{h}_t}{dt} = f(\mathbf{h}_t, \theta)
    \label{eq:2}
\end{equation}
where, $f$ is a neural network parameterized by  $\theta$ that defines the ODE dynamics. $\mathbf{h}_t$ is the hidden state of the Neural ODE. Thus, starting from an initial point $\mathbf{h}_0$, the transformed state at any time $t_i$ is given by integrating an ODE forward in time, given as,
\begin{equation}
\begin{aligned}
\mathbf{h}_i = \mathbf{h}_0 + \int_{t_0}^{t_i} \frac{d\mathbf{h}_t}{dt} dt \\
\mathbf{h}_i = ODESolve (\mathbf{h}_0, (t_0, t_i), f)
\end{aligned}
    \label{eq:3}
\end{equation}
\eqref{eq:3} can be solved numerically using any ODE Solver (e.g., Euler's method). In order to train the parameters of the ODE function $f$, an adjoint sensitivity approach is proposed in \cite{chen2018neural}. This approach computes the derivatives of the loss function with respect to the model parameters $\theta$ by solving a second augmented ODE backward in time. 
Some of the advantages of using Neural ODE solvers over other conventional approaches are: (1) \textit{Memory efficiency:} The adjoint sensitivity approach allows us to train the model with constant memory cost independent of the layers in the ODE function $f$;
(2) \textit{Adaptive computation:} Adaptive number of updation steps by ODE solver is more effective than fixed number of steps in vanilla DNN.
(3) \textit{Effective formulation:} Finally, the 
continuous dynamics of the hidden states can naturally incorporate data which arrives at arbitrary times. 

Based on the foundations of NeuODEs, ODE-RNN was developed. ODE-RNN is autoregressive in nature unlike the generative form of NeuODES \cite{rubanova2019latent}. 
It models the irregularly sampled time series by applying ODE and RNN interchangeably through a time series. The states evolve continuously using the ODE model, while they are updated using the RNN cell at the instances where the measurements are available. The function $f$ at time $t$ is described in a neural  ODE with initial hidden state $h_t$:
\begin{equation}
    \dot{\mathbf{h}} = f(\mathbf{h}_t, \theta)
\end{equation}
\begin{equation}
    o_t = c(\mathbf{h}_t, \theta_c)
\end{equation}
where, $\mathbf{h} \in \mathbb{R}^m$ is the hidden state of the data,  $\dot{\mathbf{h}} = \frac{d\mathbf{h}}{dt}$ is the derivative of the hidden state, $o \in \mathbb{R}^d$ is the output of the ODE-RNN. $f:\mathbb{R}^m \to \mathbb{R}^m $ and $c:\mathbb{R}^m \to \mathbb{R}^d$ are the neural ODE operator and output function parameterized by neural network parameters $\theta$ and $\theta_c$, respectively. The hidden states are modeled using a Neural ODE, where they obey the solution of an ODE as, 
\begin{equation}
    \mathbf{h}_{i}^{'} = ODESolve (\mathbf{h}_{i-1}, (t_{i-1}, t_{i}), f)
\end{equation}
Lets represent the RNN cell function by $v(\cdot)$ with parameters $\theta_v$. 
At each observation $x_i$, the hidden states are updated by an RNN as,
\begin{equation}
   {\mathbf{h}_{i}}  = v(\mathbf{h}_{i}^{'},x_i, \theta_v).
   \label{rnncell}
\end{equation}
The ODE-RNN approach proposed in \cite{rubanova2019latent} provides imputations but fails to quantify the uncertainty associated with it. Therefore, we offer a novel SDE-RNN framework, which, in addition to the predictions, is capable of quantifying the uncertainties in a principled manner.

\section{Proposed SDE-RNN Approach}
In this section, we describe the key elements of the proposed SDE-RNN approach. Our approach involves the modification of  ODE-RNN to include a stochastic differential equation (SDE-RNN) to capture the evolution of the hidden states (as opposed to ODE). Stochastic differential equation that governs the dynamics of the hidden states corresponds to, 
\begin{equation}
   d\mathbf{h}_t = f(\mathbf{h}_t, t) dt + g(\mathbf{h}_t, t) d\mathbf{B}_t
    \label{sde}
\end{equation}
where, $f$ and $g$ are the drift and diffusion functions, respectively. The transformations $f$ and $g$ are carried through neural networks. $\mathbf{B}_t$ represents the Brownian motion with the distribution $ d\mathbf{B}_t \sim  \mathcal{N}(0, \mathbf{Q} \Delta t)$ where $\mathbf{Q}$ is the diffusion matrix. $g(\mathbf{h}_t, t)$ denotes the variance of the Brownian motion and represents the epistemic uncertainty in the hidden state dynamics. Both the drift and diffusion functions are nonlinear, thanks to the nonlinear transformation of neural networks $f$ and $g$. Therefore, the SDEs in (\ref{sde}) is a nonlinear SDE. The hidden states in the SDE-RNN model are the solution of the SDE in (\ref{sde}), i.e.,
\begin{equation}
    \mathbf{h}_{i}^{'} = SDESolve (\mathbf{h}_{i-1}, (t_{i-1}, t_{i}), f, g)
\end{equation}
Here, SDESolve represents the SDE Solver. 
It is important to note that states are updated based on observations $x_i$ at time $t_i$ using the RNN model.
 Using this SDE-RNN approach, we can quantify the uncertainty in the hidden and output states, as discussed in the following subsection.  
\begin{figure*}[h!]
	\centering
	\includegraphics[width=\textwidth]{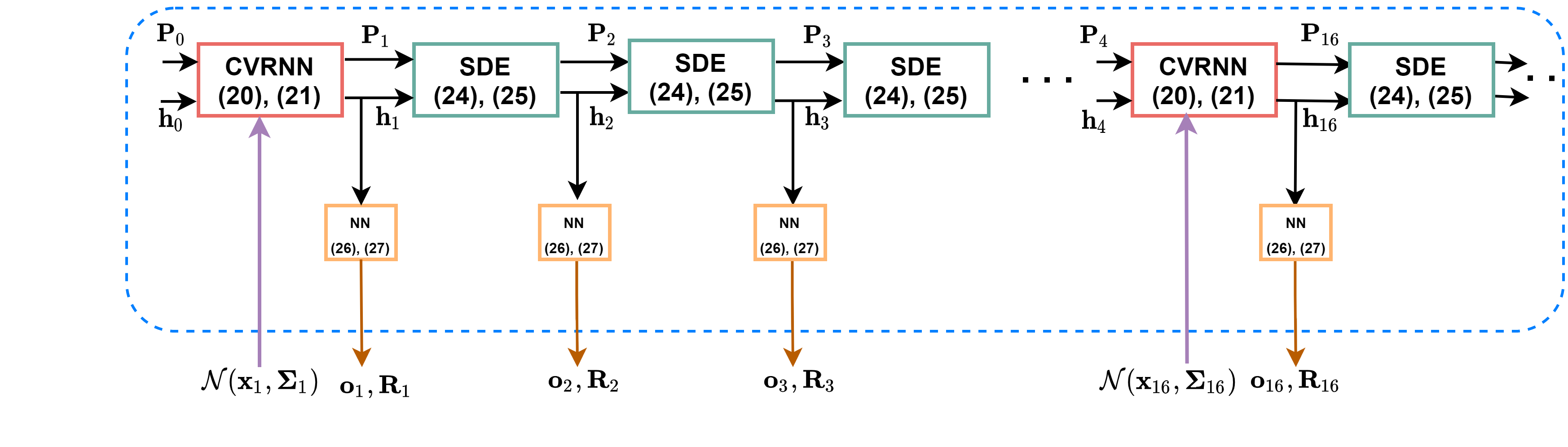}
	\caption{Propagation of uncertainty. $x_{i}$: Input at irregular intervals}
	\label{fig:framework1}
\end{figure*}
\subsection{Uncertainty quantification}
We propose a novel uncertainty propagation approach in the SDE-RNN model. The proposed method quantifies the uncertainty in a holistic manner accounting for both aleatoric and epistemic, without relying on prior specification of model parameters and complicated Bayesian inference. The SDE-RNN approach propagates the uncertainty from the noisy input observations to the final outputs using the CVRNN model and the SDE model. Here, we refer to the CVRNN model as a modified form of RNN, that propagates the mean and uncertainty arising from the previous hidden states and the inputs to the next hidden state. Fig. \ref{fig:framework1} and Algorithm  \ref{SDERNN_approach} illustrate the overall uncertainty propagation in the SDE-RNN framework. We can initialize the mean and covariance of the hidden state as zero and then update them continuously. The CVRNN model updates the hidden states at the time when an observation is available  along with its associated uncertainty.
This approach accounts for the uncertainty in both the input observation and the previous hidden states. SDE model captures the epistemic uncertainty of the SDE-RNN approach. In the following subsections, we elaborate the uncertainty propagation in both the CVRNN model and SDE model.

\subsubsection{CVRNN Model}
The RNN cell in Eq. (\ref{rnncell}) updates the hidden states at time instants where measurements are available. The dynamics of a general RNN cell are modeled by,
\begin{equation}
    {\mathbf{h}_{i}} = v(\mathbf{h}_{i-1}, x_i, \theta_v),
\end{equation}
\begin{equation}
        o_i = c(\mathbf{h}_i, \theta_c).
\end{equation}
Assume the input $x_i$ is corrupted with additive Gaussian noise given as
\begin{equation}
    \Tilde{x}_i = {x}_i + w_i,
\end{equation}
where $w_i$ = $\mathcal{N}(0, \Sigma_i$) with  $\Sigma_i$ representing the covariance matrix of the measurement.
These noisy observations are fed to the RNN model, where the hidden states and the outputs become a random variable corresponding to,
\begin{equation}
    {\Tilde{\mathbf{h}}_i} = v(\Tilde{\mathbf{h}}_{i-1}, \Tilde{x}_i , \theta_v),
    \label{eq8}
\end{equation}
\begin{equation}
        \Tilde{o}_i = c(\Tilde{\mathbf{h}}_i, \theta_c).
\end{equation}

Let $\hat{\mathbf{h}}_i = \operatorname{\mathbb{E}}[\Tilde{\mathbf{h}_i}]$ and $\hat{P_i}$ denotes the estimate of the covariance matrix $P_i = \operatorname{\mathbb{E}} \{(\Tilde{\mathbf{h}_i} - \hat{\mathbf{h}_i})(\Tilde{\mathbf{h}_i} - \hat{\mathbf{h}_i})^\intercal \}$.
Based on (\ref{eq8}), the linearization around $\hat{\mathbf{h}}_{i-1}$ implies that,

\begin{equation} \label{eq1}
\begin{split}
 {\Tilde{\mathbf{\mathbf{h}}}_i} & = v(\Tilde{\mathbf{h}}_{i-1}, \Tilde{x}_i) \\
 & = v(\hat{\mathbf{h}}_{i-1} + \delta h_{i-1}, \hat{x}_i + w_i) \\
 & =  v(\hat{\mathbf{h}}_{i-1}, \hat{x}_i)  + \Delta_h v \delta \mathbf{h}_{i-1} + \Delta_x v w_i + o(\delta \mathbf{h}_{i-1}^2, w_i^2)
\end{split}
\end{equation}
where, $\Delta_h v$ and $\Delta_x v$ are calculated at operating points $(\hat{\mathbf{h}}_{i-1},x_i)$. 
According to the transformation of uncertainty \cite{amini2021robust} given in Theorem 1, we can evaluate the expected value and covariance of $\Tilde{\mathbf{h}}_{i}$. 




\begin{theorem}
   \textit{Transformation of uncertainty:} Consider a recurrent neural network represented by $v(\cdot)$ with neural network parameters $\theta_v$ and hidden states given by  ${\Tilde{\mathbf{h}}_i} = v(\Tilde{\mathbf{h}}_{i-1}, \Tilde{x}_i , \theta_v)$. The input  $\Tilde{x}_i$ is corrupted with noise $w_i$ = $\mathcal{N}(0, \Sigma_i)$. The estimation of the expected value  and covariance matrix of the hidden state ${\Tilde{\mathbf{h}}_i}$  at time $i$  can be recursively calculated as,
    \begin{equation}
    \hat{\mathbf{h}}_i = v(\hat{\mathbf{h}}_{i-1}, x_i)
\label{hiddenstates_uncertainty}
\end{equation}
\begin{equation}
    \hat{\mathbf{P}}_i = (\Delta_h v) \hat{\mathbf{P}}_{i-1} (\Delta_h v)^\intercal +  (\Delta_x v) \Sigma_i (\Delta_x v)^\intercal
    \label{hiddencovariance_uncertainty}
\end{equation}
\end{theorem}
where, the expected value of the previous hidden state  $ \operatorname{\mathbb{E}}[\Tilde{\mathbf{h}}_{i-1}] = \hat{\mathbf{h}}_{i-1} $ and its associated covariance matrix is ${\hat{\mathbf{P}}_{i-1}}$ are given.

Thus, the mean and covariance can be computed in a CVRNN framework with noisy input measurement data at time $i$  using (\ref{hiddenstates_uncertainty}) and (\ref{hiddencovariance_uncertainty}).  We discuss the SDE model in the following subsection.

\subsubsection{SDE Model}
A neural SDE model aims to capture the epistemic uncertainty with Brownian motion and propagate it to the next time step. It also aims to capture the uncertainties arising from the previously obtained hidden states and their co-variances. A neural SDE can be expressed using the dynamical equation in Eq. (\ref{sde}).  
As this SDE is nonlinear, its statistics can be computed by adopting certain approximations. Linearized approximation of SDEs \cite{sarkka2019applied}
is a technique which computes the statistics by linearizing the drift and diffusion function around a certain point. Using the Taylor series, the drift function $f(\mathbf{h},t)$ around the mean $\mathbf{m}$ is linearized as, 
\begin{equation}
    f(\mathbf{h},t) \approx f(\mathbf{m},t) + \mathbf{F}_h(\mathbf{h},t) (\mathbf{h}-\mathbf{m})
\end{equation}
The diffusion function is linearized as,
\begin{equation}
    g(\mathbf{h},t) \approx g(\mathbf{m},t) + \mathbf{G}_h(\mathbf{h},t) (\mathbf{h}-\mathbf{m})
\end{equation}

A linearized approximation to estimate the mean and covariance of an SDE  (\ref{sde}) can be obtained by integrating the following differential equations from the initial conditions  $\mathbf{m}(t_0) = E[\mathbf{h}(t_0)]$ and $\mathbf{P}(t_0)$ = 
 Cov$[\mathbf{h}(t_0)] $  to the target time t:
  \begin{equation}
     \frac{d\mathbf{m}}{dt}  = f(\mathbf{m},t),
     \label{sdemean}
 \end{equation}
  \begin{equation}
     \frac{d\mathbf{P}}{dt} = \mathbf{P} \mathbf{F}_h^\intercal (\mathbf{m},t) + \mathbf{F}_h(\mathbf{m},t)\mathbf{P} + \mathbf{G}_h(\mathbf{m},t) \mathbf{Q} \mathbf{G}_h^\intercal(\mathbf{m},t)
     \label{sdecovariance}
 \end{equation}

Thus, (\ref{sdemean}) and (\ref{sdecovariance}) captures the evolution of both the mean and covariance of the hidden states.  
After the hidden states are updated, they are transformed by another neural network to get final predictions (see Fig.\ref{fig:framework1}). The estimation of the mean and the covariance of the output is obtained according to the transformation of uncertainty (Theorem 1) as,
     \begin{equation}
    \hat{o}_i = c(\hat{\mathbf{h}}_{i})
\label{outputstates_uncertainty}
\end{equation}
\begin{equation}
    \hat{R}_i = (\Delta_h c) \hat{\mathbf{P}}_{i} (\Delta_h c)^\intercal 
    \label{outputcovariance_uncertainty}
\end{equation}

Algorithm \ref{SDERNN_approach} illustrates the working mechanism of the proposed SDE-RNN approach. As discussed earlier, the hidden states $\mathbf{m}_i$ are updated at times $i$ where measurements are present. The availability of the measurements is indicated by the vector mask $\in \{0,1\}$. The imputed measurements along-with the associated  variances are indicated by $\{o_i\}_{i=0}^N$ and 
$\{R_i\}_{i=0}^N$ respectively. 

\begin{algorithm}
\KwInput{Datapoints $\{ x_i\}_{i=1}^{N}$ and the corresponding times $\{ t_i\}_{i=1}^{N}$, drift function 
 $f$, diffusion function  $g$\\}
Initialization: $\mathbf{h}_{0}$, $\mathbf{P}_{0}$  
\begin{algorithmic}[1]

\FOR{$i = 1,..,N$}
\STATE ${\mathbf{h}_{i}^{'}, \mathbf{P}_{i}^{'}} = SDESolve(\mathbf{h}_{i-1}, \mathbf{P}_{i-1},  (t_{i-1}, t_{i}), f, g)$ 

\STATE ${\mathbf{h}_{i}, \mathbf{P}_{i}} = CVRNN(\mathbf{h}_{i}^{'},\mathbf{P}_{i}^{'}, x_i)$ 

\STATE $\mathbf{h}_{i}$ = mask $\times$ $\mathbf{h}_{i}$ + (1-mask) $\times$ $\mathbf{h}_{i}^{'}$

\STATE $\mathbf{P}_{i}$ = mask $\times$ $\mathbf{P}_{i}$ + (1-mask) $\times$ $\mathbf{P}_{i}^{'}$

\STATE $\{o_{i}, R_{i}\}_{i=1}^{N} = c(\mathbf{h}_{i} ,\mathbf{P}_{i} )$
\ENDFOR
    \STATE \textbf{return} $\{o_i\}_{i=0}^N$, 
$\{R_i\}_{i=0}^N$
  \end{algorithmic}
  \caption{SDE-RNN Approach}
  \label{SDERNN_approach}
\end{algorithm}

\subsection{Uncertainty propagation for GRU}
The CVRNN model discussed in the previous section can be any form of the recurrent neural network; for example, it could be either RNN, Long Short Term Memory (LSTM), or Gated Recurrent Unit (GRU) model whose hidden states are given as,  
\begin{equation}
    {h_{t}} = v(h_{t-1}, x_t, \theta_v),
\end{equation} 
In this section, we will derive the uncertainty propagation for a GRU model by using the transformation of uncertainty as discussed in Theorem 1. To begin with, we need to compute the $\Delta_h v$ and $\Delta_x v$. In a GRU model, the hidden states are given as,
\begin{equation}
    h_t = z_t \circ h_{t-1} + (1-z_t) \circ h_{t}^{'} 
    \label{hidden_gru}
\end{equation}
\begin{equation}
    h_{t}^{'} = \text{tanh}(W_{in} x_t + b_{in} + r_t(W_{hn} h_{t-1} + b_{hn}))
\end{equation}
\begin{equation}
    z_t = \sigma (W_{iz} x_t + b_{iz} + W_{hz} h_{t-1} + b_{hz})
\end{equation}
\begin{equation}
    r_t = \sigma (W_{ir} x_t + b_{ir} + W_{hr} h_{t-1} + b_{hr})
\end{equation}
Here, $x_t$, $z_t$ and $r_t$ are the inputs, update and reset gates of the GRU model, respectively. \text{tanh} and $\sigma$ are the Tanh and Sigmoid activation functions, respectively. $\circ$ represents the Hadamard product. 

The gradient of the hidden states (\ref{hidden_gru}) at time $t$ computed at operating points  $(\hat{h_{t-1}}, x_t)$ are given as,
\begin{equation}
    \Delta_h v = \frac{\partial z_t}{\partial h_{t-1}} h_{t-1} + z_t  \frac{\partial h_{t-1}}{\partial h_{t-1}} + \frac{\partial (1-z_t)}{\partial h_{t-1}} h_{t}^{'} + \frac{\partial  h_{t}^{'} }{\partial h_{t-1}} (1-z_t)
    \label{gradient_eq}
\end{equation}

The derivative of the sigmoid function is denoted as $ \sigma^{'} (x) =  \sigma (x) (1- \sigma (x))$. Each gradient in the (\ref{gradient_eq}) is given as,
\begin{equation}
    \frac{\partial z_t}{\partial h_{t-1}} = \sigma^{'} (W_{iz} x_t + b_{iz} + W_{hz} h_{t-1} + b_{hz}) \circ W_{hz} 
\end{equation}
 \begin{equation}
    \frac{\partial r_t}{\partial h_{t-1}} = diag(r_t (1-r_t)) \circ W_{hr}
\end{equation}

\begin{equation}
    \frac{\partial  h_{t}^{'} }{\partial h_{t-1}} = diag(1-h_t^{'2}) \circ \{\frac{\partial r_t}{\partial h_{t-1}} (W_{hn} h_{t-1} + b_{hn}) +W_{hn} r_t \}
\end{equation}


Similarly,  $\Delta_x v$ is given as,

\begin{equation}
     \Delta_x v = \frac{\partial z_t}{\partial x_{t}} h_{t-1} + \frac{\partial (1-z_t)}{\partial x_{t}} h_{t}^{'} + \frac{\partial  h_{t}^{'}}{\partial x_{t}} (1-z_t)
     \label{gradient_eq2}
\end{equation}
Each gradient in (\ref{gradient_eq2}) is given as,
\begin{equation}
     \frac{\partial z_t}{\partial x_{t}} = diag(z_t (1-z_t)) \circ W_{iz}
\end{equation}
\begin{equation}
     \frac{\partial r_t}{\partial x_{t}} = diag(r_t (1-r_t)) \circ W_{ir}
\end{equation}
\begin{equation}
     \frac{\partial  h_{t}^{'}}{\partial x_{t}} = diag((1-h_t^{'2}) \circ (W_{in} + \frac{\partial r_t}{\partial x_{t}} (W_{hn} h_{t-1} + b_{hn}))
\end{equation}
Using Eq. (\ref{gradient_eq}) and (\ref{gradient_eq2}), the covariance matrix of the hidden states $P_t$ at time $t$ is given by (\ref{hiddencovariance_uncertainty}).
\section{Experimental results}
We validated the proposed SDE-RNN approach for imputations on irregularly sampled measurements from the power distribution network (IEEE  37 bus test system). It is important to note that the proposed framework can be applied to any irregularly sampled time series data. 
\subsection{Data pre-processing}

The irregularly sampled measurements are considered from two sensors, namely the smart meters and SCADA sensors. 
The smart meter measurements consist of active and reactive power injection for 24-hrs duration. This 24-hr load profile consists of a mixture of load profiles, i.e., industrial/commercial load profiles \cite{carmona2013fast}, and residential loads \cite{al2016state}. Reactive power profiles are obtained by assuming a power factor of $0.9$ lagging. The SCADA measurements are the voltage magnitude measurements obtained by executing load flows on the test network. The SCADA measurements are assumed at a subset of node locations. The aggregated smart meter data are averaged over 15-minute intervals while the SCADA measurements are sampled at 1-min interval. Gaussian noise with $0$ mean 
and standard deviation equal to $10\%$ of the actual power values is added to the smart meter data to mimic real-world measurement noise. The smart meter and SCADA measurements constitute our training dataset. 

The sensors data is represented as a list of records, where each record represents the information about the time-series data with the format given as,
\textit{record = [measurement type, values, times, mask]}. 
Here, time-series data at each node of the IEEE 37 bus network represents one \textit{record}. The \textit{measurement type} denotes the sensor type, i.e., $P,Q,$ or $V$. \textit{Values} $\in \mathbb{R}^{N \times 1}$ represent the sensor measurements with \textit{times} $\in \mathbb{R}^{N}$ as the corresponding time instants. \textit{Mask} $\in \mathbb{R}^{N \times 1}$ represents the availability of the corresponding measurements. The dataset is further normalized between [0,1] intervals and takes the union of all time points across different nodes in the dataset that are irregularly sampled.
\subsection{Model Specifications}

We use  the GRU cell with hidden size 5 to encode the observations. The  drift function of the SDE is a feedforward neural network with 1 layer and 100 units.  The diffusion function is a feedforward neural network with 1 layer, 100 units, and a sigmoid activation function.  We consider Ito SDE with diagonal noise of Brownian motion. Python is used
for coding the entire framework with the support of PyTorch’s
torchsde package \cite{li2020scalable}. We consider the output neural network as a 1-layer feedforward network with an input size of 5 and an output size of 1. We set the learning rate  to 0.01, batch
size as 10 and report the loss as mean squared error (MSE). We compare the efficacy of the proposed SDE-RNN with the classic GRU approach. The classic GRU approach consists of a GRU model with 3 input features (observations, times, and mask) and 5 output features. The predictions are obtained by transforming the output of the GRU cell via another feedforward neural network with 2 layers. The first layer is a neural network with output features 100 followed by the Tanh activation function. We then employ a dropout rate of 0.3 after this activation function followed by another feedforward neural network of input features 100 and output feature 1.   

The uncertainty estimates provided by the SDE-RNN is compared with that of  Monte-Carlo dropout approach in classic RNN using the Expected normalized calibration error (ENCE) metric \cite{levi2022evaluating}. Calibration error is typically used to evaluate uncertainty estimates since they are not associated with any groundtruth values. In a well calibrated model, average confidence score should match with the average model accuracy \cite{guo2017calibration}. Thus, in a classification setting, calibration implies that whenever a forecaster assigns a probability of $p$ to an event, that event should occur about $p\%$ of the time. On the other hand, in a regression setting, calibration signifies that the prediction $y_t$ should fall in a  $c\%$ (e.g., $90\%$) confidence interval approximately $c\%$ (e.g., $90\%$) of the time. The expected normalized calibration error (ENCE) is based on the similar regression setting that for each value of uncertainty, measured through the standard deviation $\sigma$, the expected mistake (measured in MSE) matches the predicted error $\sigma^2$, i.e.,
\begin{equation}
    E_{x,y} [(\mu(x) - y)^2| \sigma(x)^2 = \sigma^2] = \sigma^2
\end{equation}

ENCE metric is evaluated using the binning approach, where we assume that the number of bins $N$ divides by the number of time points $T$. We divide the indices of the examples to $N$ bins, $\{ B_j\}_{j=1}^N$, such that $B_j =\{(j-1)\frac{T}{N}+1,..., j\frac{T}{N}\}$. Each bin represents an interval in the standard deviation axis: $[min_{t \in B_{j}} \{\sigma_t\}, max_{t \in B_{j}} \{\sigma_t\}]$. Expected Normalized Calibration Error ($ENCE$) is calculated as,
\begin{equation}
     ENCE = \sqrt{\frac{1}{N} \sum_{j=1}^{N} \frac{|mVAR(j) - RMSE(j)|}{mVAr(j)}}
\end{equation}
where, the root of the mean variance ($mVAr (j)$) at bin $j$ is, 

\begin{equation}
    mVAr(j) = \sqrt{\frac{1}{| B_j|} \sum_{t \in B_j} \sigma_t^2},
\end{equation}
and root mean squared error ($RMSE(j)$) at bin $j$ is, 
\begin{equation}
    RMSE(j) = \sqrt{\frac{1}{| B_j|} \sum_{t \in B_j} (y_t - \hat{y}_t)^2}.
\end{equation}
It is expected that the $mVar$ will equal the $RMSE$ for each bin, i.e., the plot of  $RMSE$ as a function of $mVar$ should be an identity function. However, in reality, the models are not well calibrated and the plot is not an identity function. Thus, a model is said to provide better uncertainty estimates if they have a lower ENCE. 
We compare the uncertainty estimation of our SDE-RNN approach with the classic RNN in the following subsection.
\begin{figure}[]
	\includegraphics[width=0.55\textwidth]{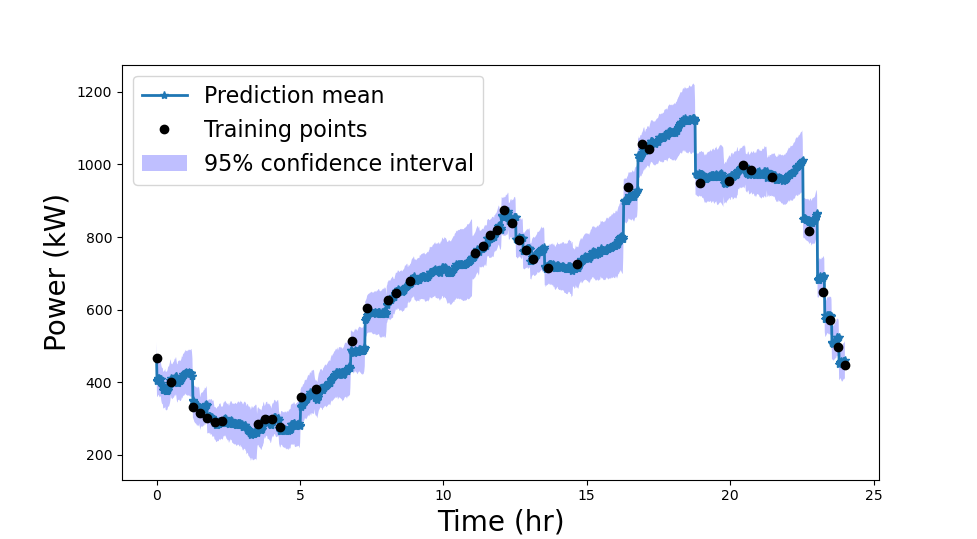}
	\caption{Uncertainty quantification in SDE-RNN approach- node 1}
	\label{fig:imputation_node1}
\end{figure}

\begin{figure}[]
	\includegraphics[width=0.55\textwidth]{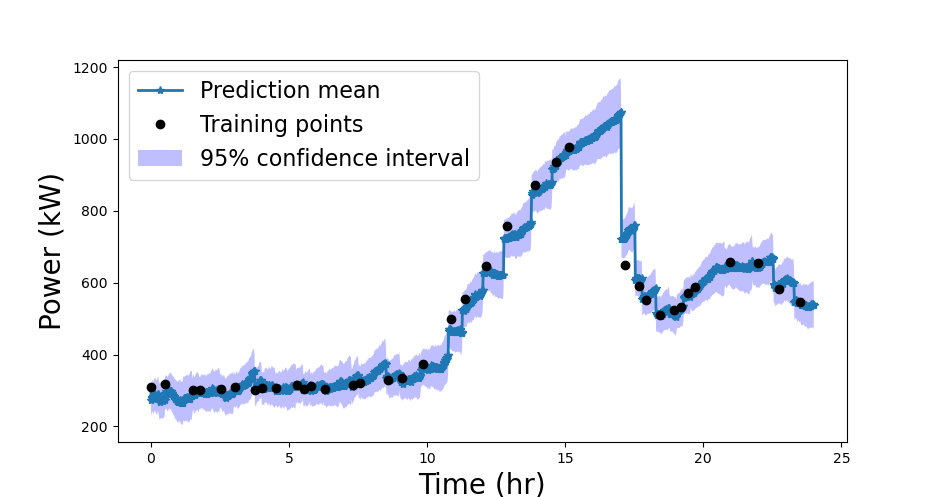}
	\caption{Uncertainty quantification in SDE-RNN approach- node 12}
	\label{fig:imputation_node12}
\end{figure}

\subsection{Imputation:}
In the first experiment, the smart meter measurements are imputed at a 1-min interval using the proposed SDE-RNN approach. The training and test dataset contains the available AMI (15-minute) and SCADA (1-minute interval) measurements. In addition, in the test dataset,  we introduce missing data as a percentage of the total number of time instants (minutes) in the 24-hr interval (i.e., 1440 time instants). The prediction and uncertainty estimates provided by the SDE-RNN model are shown in Fig.\ref{fig:imputation_node1} and Fig. \ref{fig:imputation_node12} for nodes 1 and 12, respectively. Here, $5\%$ of total interpolated points (i.e., 1440 data points) are observed. It can be observed that the epistemic uncertainty increases as we move away from the observed values. For instance in Fig.  \ref{fig:imputation_node1}, the sparse region of $10^{th}$-$12^{th}$ hr depicts less confidence compared to data-rich regions of $5^{th}$-$9^{th}$ hr. The aleatoric uncertainty due to the sensor noise is well captured by the SDE-RNN model, reflected by the uncertainty estimates at the observations. The uncertainty estimates at the observation times is nonzero and present due to the noisy input data, which is the aleatoric uncertainty part.

In the second experiment, we compare the uncertainty estimates provided by the SDE-RNN with the classic GRU + MC dropouts approach. We assume the number of bins is 5. Table \ref{imputation_table} shows the performance of both these approaches for different percentages of missing data in the test dataset. We provide the test dataset's mean squared error (MSE) and ENCE. It is evident from the Table that the SDE-RNN approach offers better accuracy and uncertainty estimates than the classic GRU model for all the levels of missingness in the test dataset. 

\begin{table}[]
\caption{Imputation results for different percentages of missing data in test dataset}
\begin{tabular}{|l|lll|lll|}
\hline
Metric                                                             & \multicolumn{3}{l|}{MSE}                                           & \multicolumn{3}{l|}{ENCE}                                          \\ \hline
\begin{tabular}[c]{@{}l@{}}Missing \\ data (\%) \end{tabular} & \multicolumn{1}{l|}{40\%}   & \multicolumn{1}{l|}{60\%}   & 80\%   & \multicolumn{1}{l|}{40\%}   & \multicolumn{1}{l|}{60\%}   & 80\%   \\ \hline
SDE-RNN                                                            & \multicolumn{1}{l|}{0.0005} & \multicolumn{1}{l|}{0.0008} & 0.0013 & \multicolumn{1}{l|}{57.45}  & \multicolumn{1}{l|}{49.11}  & 54.35  \\ \hline
Classic-GRU                                                        & \multicolumn{1}{l|}{0.0506} & \multicolumn{1}{l|}{0.0978} & 0.1527 & \multicolumn{1}{l|}{534.05} & \multicolumn{1}{l|}{419.06} & 457.92 \\ \hline
\end{tabular}
\label{imputation_table}
\end{table} 

\section{Conclusion and Future work}
This paper proposes a neural SDE-RNN framework for imputing the multi-timescale measurements and quantifying the uncertainty associated with them. We capture both the aleatoric and epistemic uncertainty and propagate it across different time instances through neural SDE-RNN modules. Simulation results (ENCE, MSE) on power distribution (IEEE 37 bus) test system validate the effectiveness of our approach. As a part of our future work, We aim to compare the epistemic uncertainty obtained through our model with that of conventional approaches such as Monte-Carlo dropouts, Deep Ensemble, etc.

\bibliographystyle{IEEEtran} 
\bibliography{ref.bib}
\end{document}